%% file: example.tex
\documentclass{article}
\usepackage{comment}
\usepackage{graphicx} % Required for inserting images
\usepackage{multirow}
\usepackage{colortbl}
\usepackage{wrapfig,lipsum,booktabs}
\usepackage{xcolor}
\usepackage{breqn}
\usepackage{skeldoc}
\usepackage{adjustbox} % 引入调整框包
\usepackage{rotating}
\usepackage{pifont}
\usepackage{xspace}
\usepackage{cleveref}

\usepackage{booktabs}
\usepackage{tabularx}
\usepackage{array}
\usepackage{caption}
\usepackage{titlesec}

\usepackage{parskip}
\crefname{figure}{Fig.}{Figs.}

\newcolumntype{Y}{>{\raggedright\arraybackslash}X}

\newcommand{\textitgray}[1]{\textcolor{gray}{\textit{#1}}}
\newcommand{\redtext}[1]{\textcolor{red}{#1}}
\newcommand{\greentext}[1]{\textcolor{green!50!black}{#1}}
\newcommand{\first}[1]{\textbf{#1}}
\newcommand{\second}[1]{\underline{#1}}
\newcommand{\third}[1]{\textit{#1}}
\newcommand{\V}[1]{\mathbf{#1}}
\newcommand{\ours}{SLAMFormer-$\infty$\xspace}

\usepackage{algorithm}
\usepackage[noend]{algpseudocode}
\usepackage[preprint]{corl_2026} % Uncomment for pre-prints (e.g., arxiv); This is like ``final'', but will remove the CORL footnote.

\title{\ours: Infinite SLAM Transformer\\ for Unbounded Frontend and Backend Processing}

\author{\vspace{-.5cm}\\
    Zhijian Fang$^*$, 
    Weicheng Zheng$^*$,
    Yijun Yuan$^*$$^\dagger$,
    Weibang Wang,
    Zhuoguang Chen,\\
    Chang Sun,
    Junhao Huang,
    Kenan Li, 
    Minghui Qin,
    Hang Zhao$^\dagger$\\
    IIIS, Tsinghua University\\
\color{blue}{\tt\small{\url{https://tsinghua-mars-lab.github.io/SLAMFormer-Infinity}}}
\\
\tt\small{\{yuanyj, hangzhao\}@mail.tsinghua.edu.cn}
    \vspace{-1cm}
}

\begin{document}
\maketitle

%===============================================================================

\begin{center}
% \vspace{-1cm}
    \includegraphics[width=0.99\linewidth]{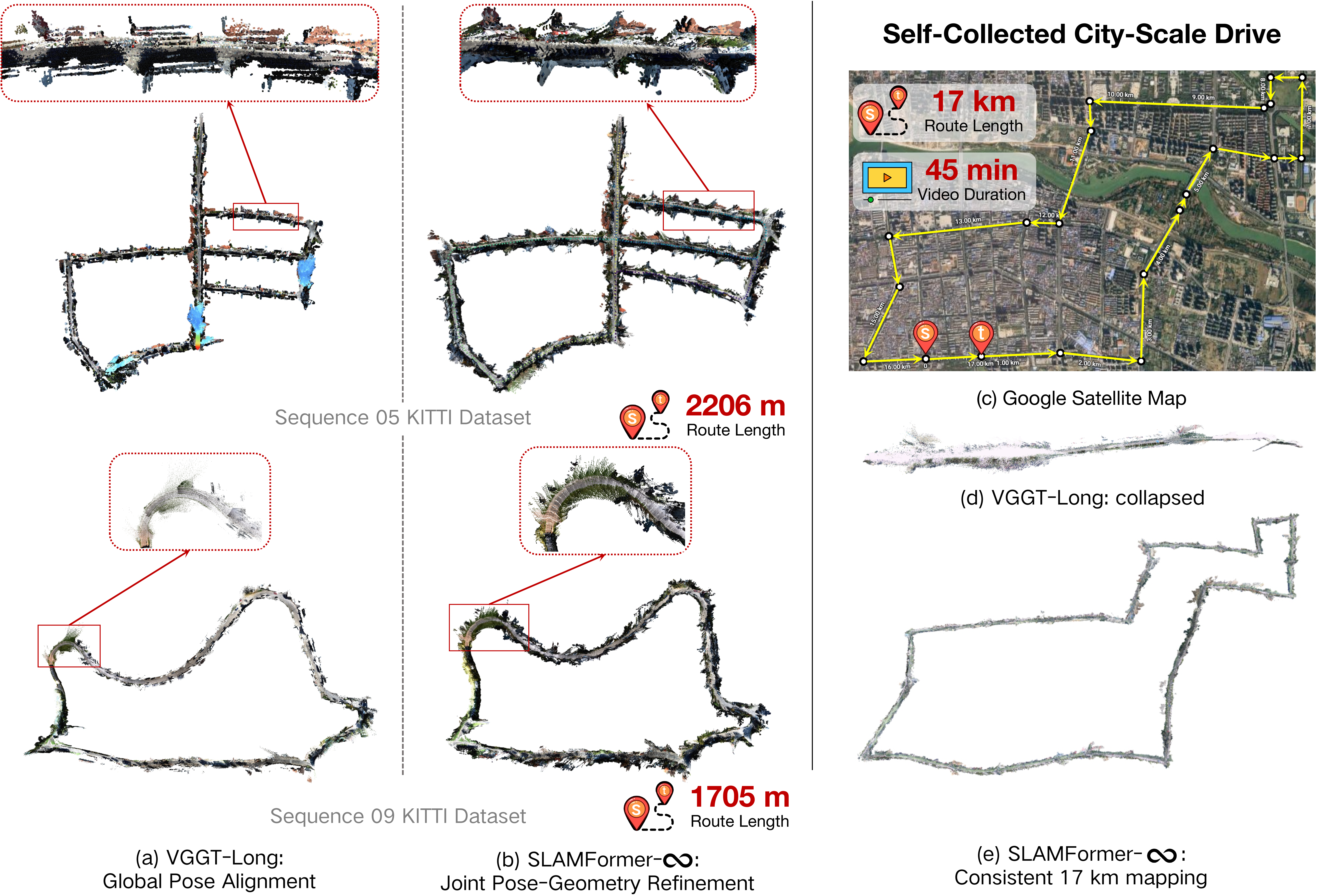}
    % \vspace{-.5cm}
    \captionsetup{type=figure}
    \captionof{figure}{\textbf{City-scale reconstruction visualization.}
VGGT-Long performs global pose alignment but leaves local geometry largely unrefined, while SLAMFormer-$\infty$ jointly optimizes pose and dense geometry.
On a self-collected 17 km urban drive, SLAMFormer-$\infty$ further maintains a consistent large-scale map where VGGT-Long collapses.}
    \label{fig:demo-long}
\end{center}
% FIXME: 1. whiteline, font, 
%        2. start-end tag. 3 zoom-in
%        3. subcaption
%        4. note 17km in figure
%        5. kitti size comparison here
%        6. 

\begin{abstract}
We introduce the Infinite SLAM Transformer (\ours), the first geometric transformer capable of supporting both long-range frontend and backend processing without an explicit distance bound.
Instead of relying on a first-frame-anchored formulation, \ours employs memory conditions to define flexible coordinate systems and scales for input frames, enabling more expressive structural conditioning.
Built upon this formulation, the frontend preserves efficient local computation, while the backend jointly optimizes long-range trajectories and scene geometry in a globally consistent manner.
Experimental results demonstrate that \ours achieves superior or highly competitive performance in both trajectory estimation and scene reconstruction across large-scale datasets.
Notably, \ours generalizes to extremely long trajectories, successfully operating on sequences exceeding $17\mathrm{km}$.
\begin{comment}
We introduce the Infinite SLAM Transformer (\ours), the first geometric transformer capable of supporting both long-range frontend and backend processing without an explicit distance bound.
\ours employs memory conditions to define customized coordinate systems and scales for input frames. Unlike previous first-frame-anchored formulations, our memory-condition design provides greater flexibility for structural conditioning.
Built upon this formulation, the frontend preserves efficient local computation, while the backend jointly optimizes long-range trajectories and scene geometry in a globally consistent fashion.
Experimental results demonstrate that \ours achieves superior or highly competitive performance in both trajectory and reconstruction across large-scale datasets.
\ours is even capable to test on $17km$ and beyond. 
We introduce the Memory-anchored SLAM Transformer, MaST, the first geometric transformer that can support both long frontend and backend without a distance bound.
% transformer
MaST tasks memory-anchor for the custom coordination and scales for input frames.
Different from the previous first-frame-anchor setup, our memory-anchor brings the freedom of structural condition.
% function
Based on our memory-anchor setup, our frontend maintains local computation; while more importantly, our backend optimizes both the long trajectory and the geometry in a global feation.
% results
From the experiment, MaST demonstrates better or highly competitive performance on both trajecotry and reconstruction on large-scale datasets. 
\end{comment}
\end{abstract}


%===============================================================================

\setlength{\abovedisplayskip}{5pt}
\setlength{\belowdisplayskip}{5pt}
\setlength{\abovedisplayshortskip}{5pt}
\setlength{\belowdisplayshortskip}{5pt}

\input{tex/introduction}

%===============================================================================

\input{tex/related_work.tex}

\input{tex/methods.tex}
\input{tex/experiments}

\section{Limitations}
%\ours still has limitations.
Unlike SLAM-Former that implicitly construct the frame-connections, \ours's PGGO inputs a pre-defined graph, from frontend and loop-detection.
The quality of graph-connectivity has effect to the performance and is not learned from data.

%The backend optimization requires an initialization, the memory anchors pre-assume in a same rough coordinate to optimize the poses and geometries of target in-between.
%So when memory anchors are fairly far-away to each other, the poses should be pre-refind before operating trajectory-geometric optimization.
%===============================================================================

\section{Conclusion}
\label{sec:conclusion}
We propose \ours to address the SLAM Transformer's limitation that cannot learn to tackle unbounded-long distance sequences.
Based on our memory-condition design, \ours achieved efficient online dense reconstruction in frontend and joint trajectory-geometry optimization in the backend for unbounded-long sequences.
Our method has demonstrated better performance across large-scale datasets and has exhibited generalization to extremely long trajectories exceeding $17\mathrm{km}$.

%===============================================================================

%\clearpage
% The acknowledgments are automatically included only in the final and preprint versions of the paper.
%\acknowledgments{If a paper is accepted, the final camera-ready version will (and probably should) include acknowledgments. All acknowledgments go at the end of the paper, including thanks to reviewers who gave useful comments, to colleagues who contributed to the ideas, and to funding agencies and corporate sponsors that provided financial support.}

%===============================================================================

\bibliography{example}  % .bib

\clearpage
\appendix

\section{Training Details}
\label{app:training_details}

\begin{table}[!h]
\centering
\caption{Dataset composition of the indoor and outdoor training configurations.}
\label{tab:training_datasets}
\small
\setlength{\tabcolsep}{4pt}
\renewcommand{\arraystretch}{1.12}
\begin{tabularx}{\linewidth}{@{}l Y c c@{}}
\toprule
Configuration & Datasets & Clip length & Long side \\
\midrule
Indoor &
ARKitScenes~\cite{baruch2021arkitscenes}, ScanNet++~\cite{yeshwanth2023scannet++}, ScanNet~\cite{dai2017scannet}, HyperSim~\cite{roberts2021hypersim}, BlendedMVS~\cite{yao2020blendedmvs}, MegaDepth~\cite{li2018megadepth} &
12 & 518 \\
Outdoor &
VirtualKITTI2~\cite{cabon2020virtual}, TartanGround~\cite{patel2025tartanground}, HabitatHM3D~\cite{yadav2023habitat}, ARKitScenes~\cite{baruch2021arkitscenes}, ScanNet++~\cite{yeshwanth2023scannet++}, BlendedMVS~\cite{yao2020blendedmvs}, MegaDepth~\cite{li2018megadepth} &
36 & 224 \\
\bottomrule
\end{tabularx}
\end{table}

\begin{table}[!h]
\centering
\caption{Optimization hyperparameters used for both training configurations.}
\label{tab:training_hyperparams}
\begin{tabular}{l l}
\toprule
Hyperparameter & Value \\
\midrule
Batch size & 1 per GPU \\
Gradient accumulation & None \\
Optimizer & AdamW \\
Weight decay & 0.05 \\
Initial learning rate & $1\times10^{-5}$ \\
Minimum learning rate & $1\times10^{-8}$ \\
LR schedule & Cosine decay \\
Warm-up & 0.5 epochs \\
Precision & Mixed precision \\
Training epochs & 10 \\
GPUs & 48 A100 \\
Indoor training time & $\sim$1 hour / epoch \\
Outdoor training time & $\sim$2.5 hours / epoch \\
\bottomrule
\end{tabular}
\end{table}

\begin{figure}[!t]
    \centering
    \includegraphics[width=0.8\linewidth]{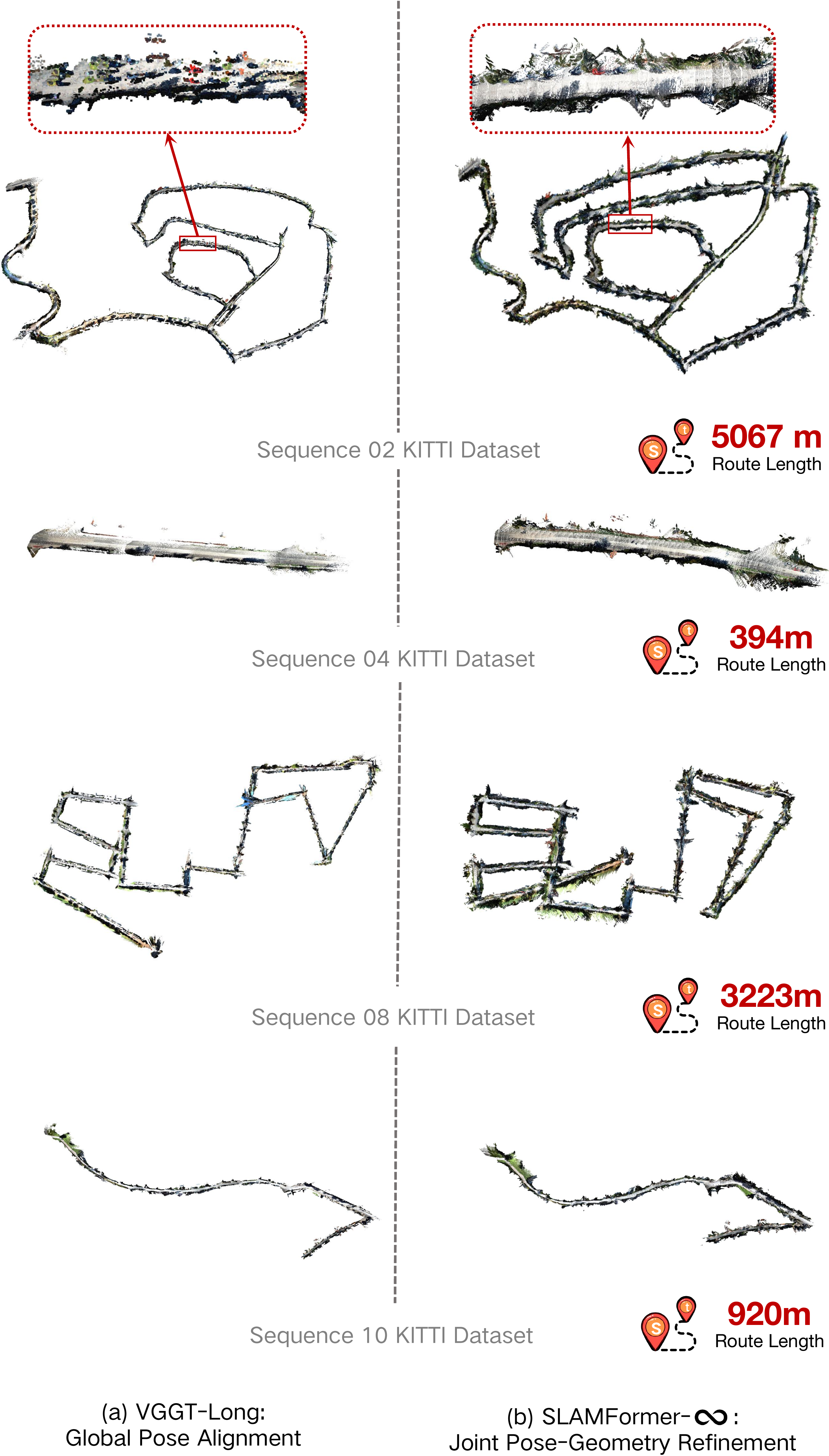}  
    \caption{
Additional qualitative comparisons on KITTI Odometry. 
VGGT-Long performs global pose alignment but leaves local geometry largely unrefined, whereas SLAMFormer-$\infty$ jointly refines pose and dense geometry, producing more coherent large-scale reconstructions.
}
    \label{fig:kitti_appendix_vis}
\end{figure}

\section{Qualitative Analysis on KITTI}
\label{app:kitti_qualitative}

Fig.~\ref{fig:kitti_appendix_vis} provides additional qualitative comparisons on KITTI Odometry sequences. 
VGGT-Long achieves global pose alignment and preserves the coarse trajectory layout, but its dense geometry is mainly stitched after pose correction. 
As a result, local structures often remain fragmented or misaligned, as shown by the zoomed regions where nearby surfaces become discontinuous and poorly fused.

In contrast, SLAMFormer-$\infty$ performs joint pose-geometry refinement through the memory-anchored backend. 
The optimized poses and pointmaps are updated together, producing more coherent local geometry while maintaining large-scale trajectory consistency over hundreds of meters to kilometer-scale routes. 
These results show that the proposed backend improves reconstruction quality beyond trajectory-level alignment, especially in long outdoor sequences where local geometric errors can accumulate after pose-only optimization.

\end{document}

%% file: tex/introduction.tex
\section{Introduction}

% SLAM's importance
%Simultaneous Localization and Mapping (SLAM) enables robot to localize itself while building maps of unknown environments.
Simultaneous Localization and Mapping (SLAM) plays a fundamental role in autonomous systems by enabling robot to localize itself and construct maps of previously unknown environments. 
%In robotic autonomous systems, SLAM plays an vital role because it enables simultaneously localize themselves within an unknown environment and construct a map of that environment, forming the basis for any intelligent spatial decision-making.
% mono slam
Among different SLAM paradigms, monocular SLAM is particularly important for its simplicity and low hardware cost, requiring only a single RGB camera for autonomous perception and navigation.
%Among different SLAM paradigms, monocular SLAM is particularly attractive due to its simplicity and low hardware cost, relying only on a single RGB camera for autonomous perception and navigation.
%Following up on the general importance of SLAM, the specific value of Monocular SLAM (MonoSLAM) lies in its unparalleled simplicity, accessibility, and low cost. It solves the autonomy problem using the most minimal and ubiquitous sensor possible: a single standard camera.
%This makes it the ``democratizing force" in robotics, capable of bringing advanced navigation to devices where size, weight, power, or budget are the primary constraints.

% the SLAM's development
% from traj to dense
Early monocular SLAM systems~\cite{davison2007monoslam,klein2007parallel}, primarily focused on camera trajectory estimation with sparse feature-based maps. Although effective for localization, these approaches captured only limited scene geometry. To recover dense three-dimensional structure, later methods introduced dense monocular SLAM through dense bundle adjustment~\cite{teed2021droid} or learning-based depth prediction~\cite{ummenhofer2017demon}.

%Early monocular SLAM systems, such as MonoSLAM and PTAM, focused predominantly on estimating the camera's six-degree-of-freedom trajectory while constructing only a sparse map of distinctive feature points. This sparse representation proved sufficient for localization tasks but inherently discarded the majority of geometric information present in the environment, leaving walls, obstacles, and surface topologies largely unknown. The subsequent paradigm shift toward dense monocular SLAM sought to recover complete pixel-level depth maps, enabling full three-dimensional surface reconstruction from a single moving camera. This transition was initially advanced by direct methods that operated on image intensity gradients rather than sparse features, and later accelerated by learning-based depth prediction networks that provided geometric priors to resolve the fundamental ill-posedness of monocular depth estimation.

% to neural 
More recently, neural scene representations and geometric foundation models have significantly advanced monocular SLAM. Gaussian Splatting-based methods~\cite{matsuki2024gaussian,zhang2025hi}
demonstrate that neural rendering representations enable high-quality dense reconstruction from monocular input. Meanwhile, Geometric foundation-based methods~\cite{murai2025mast3r,maggio2026vggt} leverage geometric transformers to predict camera pose, and scene geometry from multi-view observations, followed by backend pose-optimization for global consistency. Taking one step further, SLAM-Former~\cite{yuan2025slam} introduces transformer-based global attention for trajectory-geometry refinement without loop detection and optimization. Together, these methods move dense monocular SLAM toward learned, globally consistent, and geometry-aware systems.
%The integration of neural representations has constituted the most transformative recent advance in dense monocular SLAM. MonoGS pioneered the use of three-dimensional Gaussian Splatting for monocular SLAM, demonstrating that representing the map as millions of semi-transparent Gaussians enables real-time dense reconstruction and rendering from a single camera. HI-SLAM2 extended this paradigm by introducing geometry-aware rendering techniques, achieving RGB-D level quality using only RGB input while improving localization accuracy through better geometric consistency. Concurrently, the emergence of geometric foundation models has fundamentally reshaped the front-end of monocular SLAM. MASt3R-SLAM leverages the DUSt3R family of networks to directly predict dense point maps from image pairs, bypassing traditional feature extraction and matching. VGGT-SLAM builds upon visual geometry guided transformers, employing a fully transformer-based architecture to jointly estimate depth, camera pose, and scene geometry in an end-to-end manner. Complementing these approaches, SLAM-Former introduces a transformer-based module that replaces conventional sliding-window optimization with global graph attention over keyframes, enabling long-term loop closure and drift correction in challenging low-texture environments. Together, these methods represent a convergence of neural scene representation and geometric deep learning, moving monocular SLAM from hand-crafted pipelines toward learned, differentiable, and globally consistent systems.

\begin{figure}[!t]
    \centering
    \includegraphics[width=1\linewidth]{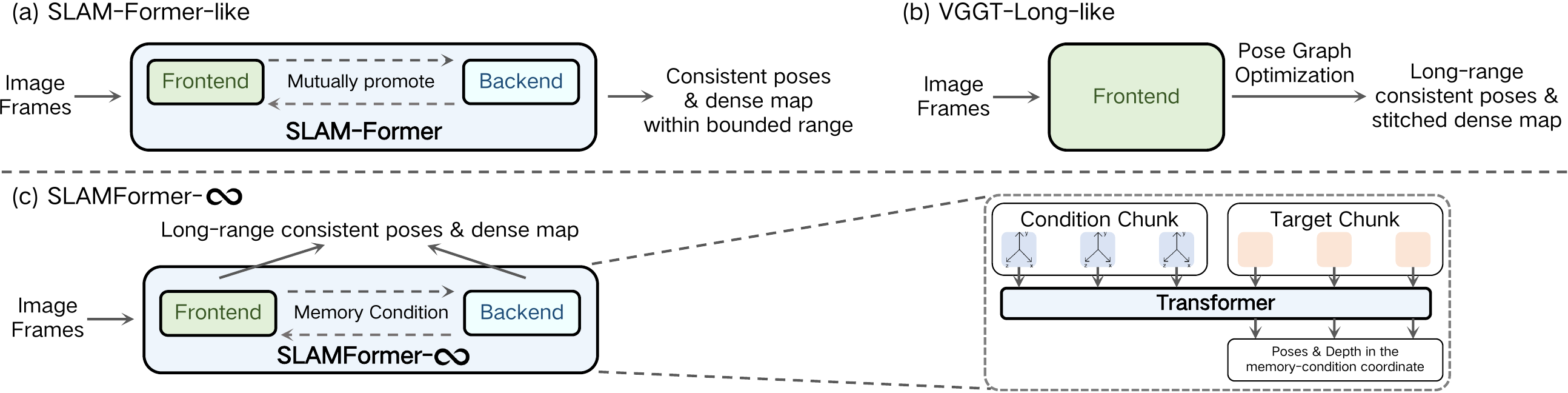}
    \caption{\textbf{SLAM Transformer Comparison.} (a) Single-model SLAM-Former for global consistent pose and map. (b) VGGT-Long with optimized long-range poses and stitched maps.
    (c) Ours retains the single-model while obtains long-range global consistent pose and map with a conditional design. 
    %that the condition chunk provides the target with reference coordinate and scale information.
    }
    \label{fig:RL}
   
    %\vspace{-.3cm}
\end{figure}
% FIXED: 1. right side, rm the middle dash line [done]
%        2. remove the global before attention. [done]
%        3. pose-condition -> Memory-condition [done]
%        4. 图片左侧体现我们解决了两个limitation的方案对比。（a）in limited range (做不长), (b) only pose in optim the vggt-like method （没有全局优化）。(c) = a+b, (d) 现在的右下角 [done]
%        

% introduce of their limitation
However, recent transformer-based monocular SLAM systems suffer from two fundamental limitations. 
On one hand, fully data-driven approaches such as SLAM-Former~\cite{yuan2025slam} rely entirely on learned trajectory modeling, making their long-range performance bounded by the scale and trajectory distribution of training data. 
%Since real-world datasets cannot provide arbitrarily long and diverse sequences, these methods often struggle to generalize toward trajectories significantly beyond the training regime. 
On the other hand, systems such as MASt3R-SLAM~\cite{murai2025mast3r} and VGGT-SLAM~\cite{maggio2026vggt} address long-range consistency through pose-centric optimization, where the backend primarily refines camera poses while leaving scene geometry largely fixed. As a result, trajectory correction and geometric reconstruction remain decoupled rather than being jointly optimized within a unified framework.

%However, despite the rapid progress of recent data-driven monocular SLAM systems, such as SLAM-Former, its performance remains fundamentally constrained by training data. In practice, real-world datasets cannot provide infinitely long trajectories or exhaustive scene diversity. Long sequences with reliable geometric supervision are expensive to collect and annotate, while existing datasets often exhibit strong distributional biases toward specific environments, such as urban driving or structured indoor scenes. As a result, current learned SLAM systems still struggle to generalize toward unbounded long-range tracking and reconstruction.

%However, although the most recent data-driven monocular SLAM systems—exemplified by MASt3R-SLAM, VGGT-SLAM, and SLAM-Former—benefit substantially from data-scaling, they remain fundamentally bounded by the acquisition of training data. 
%The critical constraint is that real-world training data cannot be infinitely long nor universally exhaustive. Long trajectories necessary for learning temporal consistency and loop closure are scarce due to the practical difficulties of collecting, storing, and annotating extended camera sequences with ground-truth poses and dense depth. Furthermore, even massive datasets exhibit inherent distributional biases—such as over-representation of urban driving scenes or structured indoor environments—which directly limits the generalization of data-driven SLAM systems to unstructured, feature-poor, or dynamically changing environments. 

% introduce our method
To address these limitations, we propose \ours, an infinite SLAM Transformer for unbounded frontend and backend optimization as in~\cref{fig:RL}. 
Unlike previous transformer-based SLAM systems, 
%ours preserves conditional frames throughout transformer inference, allowing them to serve as persistent geometric conditions for coordinate and scale prediction. 
our method preserves memory condition throughout transformer inference, enabling them to act as persistent geometric anchors.
Based on this design, \ours supports an efficient local frontend for long-range tracking, while enabling iterative backend that jointly optimizes both trajectory and geometry over arbitrarily long sequences.

%Therefore, we introduce conditioned SLAM transformer to tackle this limitation. 
%Unlike previous methods, CST treat condition frames unchanged during the transformer process. 
%And thus make it able to support window frontend for efficiency, and backend iteration for long-distance geometric-trajectory-consistency.
\begin{comment}
The contribution of this paper are as following:
\begin{itemize}
    \item A novel Conditional SLAM Transformer that supports both frontend and backend functionalities,
    \item An transformer-based SLAM frontend that supports unbounded long tracking, 
    \item An iterative transformer-based SLAM backend for unbounded long geometric and trajectory consistency,
    \item Our experiments has demonstrate much better reconstruction while highly competitive performance to the SOTAs.
\end{itemize}
\end{comment}

The contributions of this paper are summarized as follows:
\begin{itemize}[nosep]
    \item We propose a novel Infinite SLAM Transformer that supports an unbounded long-range frontend tracking and backend optimization within a conditioned transformer framework.
    \item We introduce Pose-Geometry Graph Optimization (PGGO) with transformer that jointly refines long-range trajectory and scene geometry in a globally consistent manner.
    %\item We develop training strategies that enable stable learning of both frontend tracking and backend refinement capabilities within the transformer.   
    \item Experimental results demonstrate significantly improved reconstruction quality while achieving highly competitive localization performance against state-of-the-art methods.
\end{itemize}

% FIXED: 1. contribution 1 is with unbounded with application function, 2 with model function. pose-geometry optim. [done]
%        2. contribution 3 is redundant [done]

%% file: tex/related_work.tex
\section{Related Work}

\subsection{Geometric Transformer}
Recent geometric foundation models have reshaped multi-view 3D perception by replacing hand-crafted correspondence and optimization with transformer-based geometry regression~\cite{wang2024dust3r,leroy2024grounding}.
DUSt3R~\cite{wang2024dust3r} predicts dense pointmaps from image pairs, turning matching and triangulation into learned geometric regression, while MASt3R~\cite{leroy2024grounding} further strengthens pairwise 3D matching.
Beyond pairwise inference, Fast3R~\cite{yang2025fast3r}, VGGT~\cite{wang2025vggt}, and Pi3~\cite{wang2025pi} extend this paradigm to feed-forward multi-view reconstruction, with VGGT jointly predicting cameras, depth, and pointmaps from multiple views.
However, these models are still mainly formulated for bounded input sets, where all views fit into a finite attention context and the learned coordinate behavior is tied to training-time sequence ranges.

%Recent long-sequence geometric transformers relax this assumption through memories, recurrence, test-time adaptation, or chunked processing~\cite{chen2025ttt3r,cheng2026longstream,chen2026geometric}.
%For example, TTT3R~\cite{chen2025ttt3r} improves recurrent reconstruction by adapting the model state at test time, while other chunk-based systems extend to long videos through local reconstruction and pose-based stitching.
%Although these designs improve scalability, their long-range consistency is still largely maintained through state management, coordinate resets, pose propagation, or submap alignment, rather than unified transformer refinement of distant poses and geometry.
Recent long-range geometric transformers relax this assumption through memories, recurrence, test-time adaptation, or chunked processing~\cite{chen2025ttt3r,cheng2026longstream,chen2026geometric}.
For example, TTT3R~\cite{chen2025ttt3r} adapts model states during inference for recurrent reconstruction, while extend to long videos through local reconstruction and pose stitching.
However, their long-range consistency still mainly relies on state management, coordinate resets, pose propagation, or submap alignment, rather than unified transformer refinement over distant poses and geometry.

\ours takes coordinate-conditioning view of long-sequence geometric reasoning: fixed condition chunks define the local reference geometry for active-frame prediction, so the transformer can operate on a bounded context while preserving long-range pose and geometry information.

\subsection{Learning-based SLAM}
Learning-based SLAM has gradually moved from hand-crafted visual frontends to neural geometric modules.
Optical flow-based systems such as DROID-SLAM~\cite{teed2021droid} and SceneFactory~\cite{yuan2025scenefactory} construct dense structure through dense pixel matchings, while neural implicit and Gaussian-splatting SLAM methods optimize dense maps through rendering objectives, with MonoGS~\cite{matsuki2024gaussian} demonstrating real-time monocular SLAM with 3D Gaussian Splatting.
These methods still rely on explicit optimization or iterative rendering-based updates to maintain trajectory and geometry consistency.

More recent methods use geometric foundation models as feed-forward reconstruction priors.
MASt3R-SLAM~\cite{murai2025mast3r} builds dense SLAM from MASt3R matching and pointmap prediction, VGGT-SLAM~\cite{maggio2026vggt} and VGGT-Long~\cite{deng2025vggt} construct and globally align local VGGT submaps, and SLAM3R~\cite{liu2025slam3r} follows a local-clip reconstruction and registration strategy.
These methods provide strong local geometry, but global consistency is still mainly imposed by external registration, submap alignment, pose-graph optimization, or fusion.

SLAM-Former~\cite{yuan2025slam} further moves toward transformer-native SLAM by integrating frontend tracking, mapping, and backend refinement into a single model.
However, its global refinement relies on accumulated trajectory representations and historical KV states, tying long-range inference to a growing sequence-level state.
\ours follows this unified-transformer direction, but replaces growing-state refinement with pose-geometry graph optimization (PGGO), enabling iterative joint refinement of camera poses and dense geometry with the same transformer.

%% file: tex/methods.tex
\section{Methodology}
\label{sec:cond-slamformer}

\subsection{Problem Formulation}
Given a streaming monocular image sequence
$
\mathcal{I}_{1:N}=\{\V I_1,\dots,\V I_N\},
$
the goal of SLAM is to estimate in real time both
the camera trajectory
$
\mathcal{X}_{1:N}=\{\V g_1,\dots,\V g_N\}, \quad \V g_n \in SE(3),
$
and the scene geometry representation
$
\mathcal{P}_{1:N}=\{\V P_1,\dots,\V P_N\},
$
while maintaining both accurate incremental tracking and global geometric consistency.

SLAM-Former~\cite{yuan2025slam} formulates SLAM as inference within a single transformer model:
$
p(\mathcal{X}, \mathcal{P} \mid \mathcal{I}),
$
replacing modulation with end-to-end prediction over trajectories and scene structure.

% FIXED: 1. 切换成kitti0 的展示 [done]
%        2. 字体换彩色 （black attention 不能太多）[done]
%        3. 更好展示global和local [done]
%        4. 在global backend里面加上PGGO [done]

Specifically, it consists of a frontend and backend:
\begin{align}
\text{Frontend:}\quad
&\mathcal{M}_n
=
f_\theta^{f}(\mathcal{I}_{n}, \mathcal{M}_{1:n-1}), \\
\text{Backend:}\quad
&\hat{\mathcal{M}}_{1:n}
=
f_\theta^{b}(\mathcal{M}_{1:n}).
\end{align}
where $\mathcal{M}$ denotes map token representations, and poses and geometry are decoded via a head function:
$({\mathcal{X}}, {\mathcal{P}})=f_{\psi}(\mathcal{M})$.

The frontend operates causally for incremental tracking, while the backend performs global refinement over the full history.

\subsection{\ours}
However, this formulation is fundamentally constrained by the distribution of training trajectories, preventing effective generalization in long-range sequences.

To address this limitation, we introduce Infinite SLAM transformer (\ours), modeling:
$
p(\mathcal{X}, \mathcal{P} \mid \mathcal{I}, \mathcal{I}_{C}, \mathcal{X}_{C}),
$
where the memory condition $(\mathcal{I}_{C}, \mathcal{X}_{C})$ defines a reference coordinate system.

A key distinction from SLAM-Former is that ours performs both frontend and backend inference in a local coordinate system defined by the condition, rather than in a global coordinate system.

Given a graph $\mathcal{G}=(\mathcal{V},\mathcal{E})$ over keyframes, \ours operates on local neighborhoods ($\mathcal{N}$):
\begin{align}
\label{eq:our_frontend}
\text{Conditional Frontend:}\quad
&\mathcal{M}_n
=
f_\theta^{f}(\mathcal{I}_{n}, \mathcal{I}_{n-k:n-1}, \mathcal{C}_{j \in \mathcal{N}_{(n-k)}}),\\
\text{Conditional Backend:}\quad
&\hat{\mathcal{M}}_{n-w:n}
=
f_\theta^{b}(\mathcal{I}_{n-w:n}, \mathcal{C}_{j \in \mathcal{N}_{(n-w)}}).
\label{eq:our_backend}
\end{align}
where $\mathcal{C}_\cdot$ denotes neighboring conditioning context $(\mathcal{I}_\cdot, \mathcal{X}_\cdot)$, $w\in\mathcal W(n)$ is the nearest anchor retriever out of window set $\mathcal{W}(n)$. Note that in frontend, $\mathcal{I}_{n-k:n-1}$ and $\mathcal{C}$ assists $\mathcal{I}_{n}$ with only KV caches obtained from previous backend and frontend processing.

\begin{comment}
The frontend and backends are mutually coupled through the conditional formulation. The local backend periodically refines each fixed-size chunk, truncates the active inference window, and provides updated condition information for subsequent frontend inference. Therefore, the streaming frontend is continuously optimized by backend-provided local conditions, keeping tracking and mapping under nearby in-distribution reference frames instead of extrapolating in an ever-growing global frame. The global backend further improves this process by producing globally consistent constraints after loop detection or at sequence end, which can be reused as stronger conditions for later frontend inference. In turn, the streaming frontend and local backend jointly provide the initialization for the global backend; when the accumulated error is large, we first perform pose graph optimization using the current graph constraints and use the optimized poses as the global backend initialization.
This design makes predictions always conditioned on in-distribution reference frames.

\end{comment}

However, this relative formulation prevents direct global full-attention inference as in SLAM-Former~\cite{yuan2025slam}, motivating an iterative backend for global consistency.

\subsection{Pose-Geometry Graph Optimization (PGGO)}
\label{sec:backend}
Our backend jointly processes poses and geometries.
We formulate the task as a Pose-Geometry Graph Optimization (PGGO) that jointly refines trajectory and scene geometry under long-range relational constraints.
Specifically, PGGO operates over a pose-geometry interaction graph $\mathcal{G}=(\mathcal V, \mathcal E)$, 
where node set is defined as
$\mathcal{V}=\mathcal{X}\cup\mathcal{P}$
with $\mathcal{X}$ and $\mathcal{P}$ denoting pose nodes and geometry nodes.
The edges is defined as
$\mathcal{E}=\mathcal{E}_{\mathcal{X}}\cup\mathcal{E}_{\mathcal{P}}
$, where $\mathcal{E}_{\mathcal{X}}$ connects poses explicitly with relative pose constraints, while $\mathcal{E}_{\mathcal{P}}$ implicitly captures geometric correlation through transformer attention $f_\theta$.

For each frame ($n$), we define the joint state variable as:
\(
\V x_n = (\V g_n, \V P_n)
\).
Given an initialization
\(
\tilde{\V x}^0_n = (\tilde{\V g}^0_n, \tilde{\V P}_n)
\),
obtained from pose graph optimization or frontend prediction,
our goal is to jointly refine poses and geometries for global consistency:
\begin{equation}
\V x^\ast
=
\arg\min_{\{\V x_n\}}
\sum_n
\left\|
\V x_n - f_\psi \circ f_\theta(\mathcal{I}_n, \mathcal{C}_{n\in\mathcal{N}_{n-w}})
\right\|^2.
\end{equation}

To solve above optimization, for $\V x_n\in \mathcal {V}$, we use an iterative neural refinement:
\[
\hat{\V x}_n^{k+1}
=
f_\psi \circ f_\theta(\mathcal{I}_n, \mathcal{C}_{n\in\mathcal{N}_{n-w}}),
\]
followed by direct pointmap updates together with dampled pose updates:
\begin{align}
\tilde{\V P}_n\leftarrow \hat{\V P}_n^{k+1}, \ \ \ 
\tilde{\V g}_n^{k+1}
=
\exp[(1-\alpha)\log(\tilde{\V g}_n^k) + \alpha \log(\hat{\V g}_n^{k+1})],
\end{align}
with $\alpha$ the damping factor. We write $\tilde{x}^{k+1}_n = h_{\theta}(\tilde{x}^{k}_n)$ for simplicity of PGGO iteration.

\begin{comment}
    
Let the state of node \(i\) be defined as the $x_i = (T_i, D_i)$.
%$\mathcal{M}_i$, with \((T_i,D_i)=h(\mathcal{M}_i)\).
%We initialize the graph states using the pose graph optimization (PGO) result:
%$x_i^{0} = (\tilde{T}_i, D_i^{0})$,
%where \(\tilde{T}_i\) is the globally consistent pose estimated.
Given the pose graph optimized initialization
\(
\tilde{x}_i = (\tilde{T}_i, D_i^0),
\)
our goal is to recover graph states that are both globally consistent on poses and geometries.

We formulate the refinement process as the following fixed-point optimization problem:
\[
x^\ast
=
\arg\min_{\{x_i\}}
\sum_i
\left\|
x_i -
h\circ f_\theta(\mathcal{M}_i,\mathcal{N}_i)
\right\|^2
+
\lambda
\sum_i
\left\|
T_i \ominus \tilde{T}_i
\right\|^2,
\]
where
\(
\mathcal{N}_i
=
\{x_j \mid j \in \mathcal{E}(i)\}
\)
denotes the neighboring node states,
\(\ominus\) denotes the relative pose difference operator on
\(\mathrm{SE}(3)\),
and \(\lambda\) balances local neural consistency and global pose graph consistency.

Instead of directly solving the above optimization using gradient-based methods, we employ an iterative neural relaxation process:
\[
\hat{x}_i^{k+1}
=
h\circ f_\theta(\mathcal{M}_i,\mathcal{N}_i^k),
\]
followed by a damped update:
\[
x_i^{k+1}
=
(1-\alpha)x_i^k
+
\alpha \hat{x}_i^{k+1},
\]
where \(0 < \alpha \leq 1\) is the relaxation factor.
\end{comment}

\begin{figure}[!h]
    \centering
    \includegraphics[width=\linewidth]{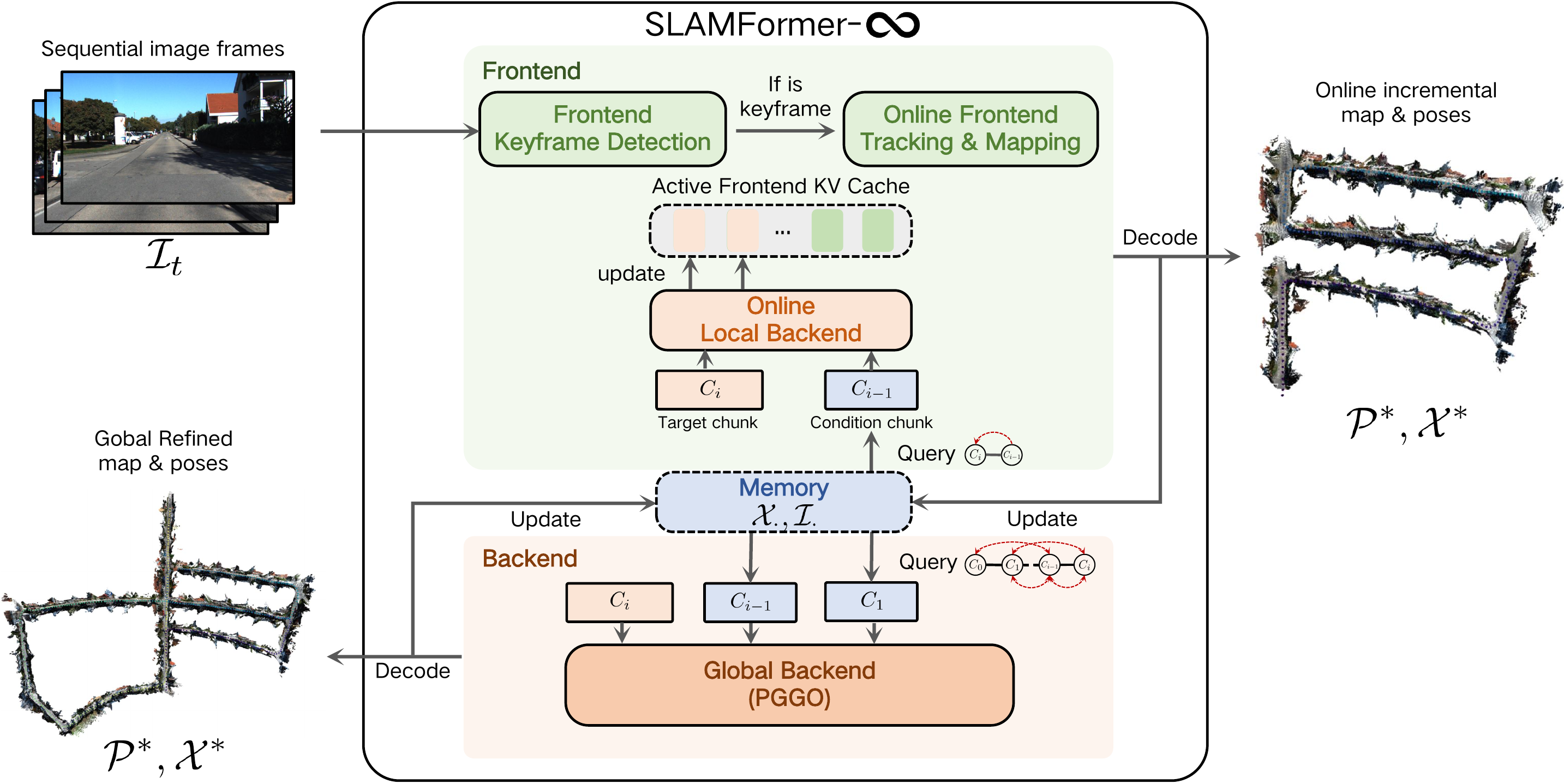}
    \caption{The SLAM pipeline. Frontend detects keyframes and provides online track. 
    After a fixed number of keyframes, we refine the most recent local window to improve short-range pose and geometry consistency.
    %each incoming frame is either associated with an existing keyframe or inserted as a new node, and new keyframes are initialized by local conditional tracking. \textbf{Local backend:} after a fixed number of keyframes, we refine the most recent local window to improve short-range pose and geometry consistency. \textbf{Global backend:} 
    When loop detection is triggered or at sequence end, PGGO is applied with the same transformer function.}
    \label{fig:demo}
        % \vspace{-.5cm}
\end{figure}

\subsection{SLAM at Test Time}

At test time as in~\cref{fig:demo}, \ours performs streaming reconstruction and pose estimation from RGB sequences $\{\V I_t\}_{t=1}^T$, with keyframes as $\{\V I_n\}_{n=1}^N$.
We maintain a global graph:
$
\mathcal{G}=(\mathcal{V},\mathcal{E}),
$
where each node $\V x_n$ stores:
$
\V x_n = (\V g_n, \V P_n).
$

\textbf{Frontend.}
Each incoming frame $\V I_i$ is either associated with an existing keyframe or inserted as a new node $\V I_n$. Edges are updated based on the local window $\mathcal{W}(n)$.
New keyframe $\V I_n$ is tracked with \cref{eq:our_frontend} for $\mathcal{M}_n$, with the assistance of previous KV caches.
%$x_n^{0} = f_\theta(I_n, M,\mathcal{N}(n))$ with the memory.

\textbf{Local Backend.}
Every $c_w$ keyframes inputs, we trigger one local backend with function eq. (4) for the update of $\hat{\mathcal{M}}_{n-c_w:n}$.

\textbf{Global Backend.}
When loop-detection is triggered or after the last frame, given the graph initialization from pose graph or frontend prediction, \ours performs PGGO for global refinement: iterating from $k=1$ to $K$, for node $\V x_n\in \mathcal {V}$, $\tilde{\V x}^{k+1}_n = h_\theta(\tilde{\V x}^{k}_n)$.
These results yield a globally consistent reconstruction induced by the learned \ours prior.

\subsection{Training Strategy}

\begin{figure*}[!t]
    \centering
    \begin{minipage}[t]{0.24\textwidth}
        \centering
        \includegraphics[width=\linewidth]{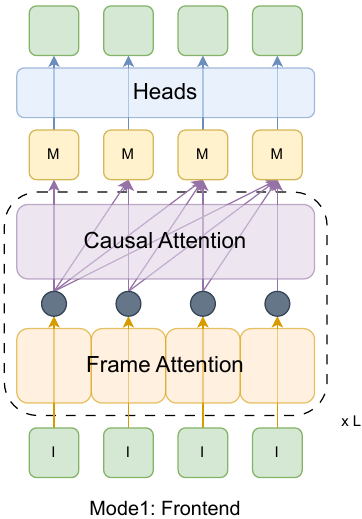}
        % \vspace{0.3em}
        % {\tiny\textbf{Mode 1: Frontend}}
    \end{minipage}
    \hfill
    \begin{minipage}[t]{0.24\textwidth}
        \centering
        \includegraphics[width=\linewidth]{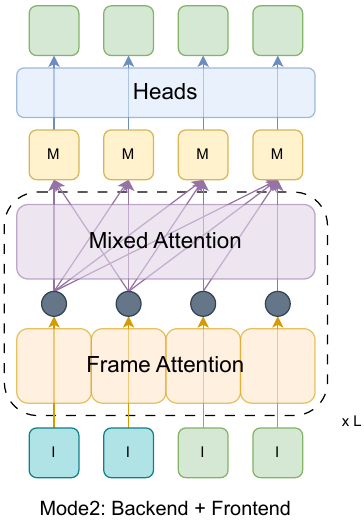}
        % \vspace{0.3em}
        % {\tiny\textbf{Mode 2: Backend + Frontend}}
    \end{minipage}
    \hfill
    \begin{minipage}[t]{0.24\textwidth}
        \centering
        \includegraphics[width=\linewidth]{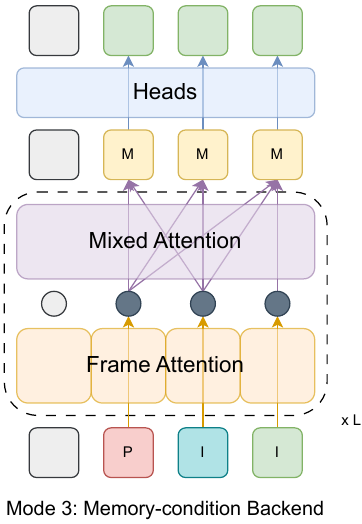}
        % \vspace{0.3em}
        % {\tiny\textbf{Mode 3: Memory-anchor Backend}}
    \end{minipage}
    \hfill
    \begin{minipage}[t]{0.24\textwidth}
        \centering
        \includegraphics[width=\linewidth]{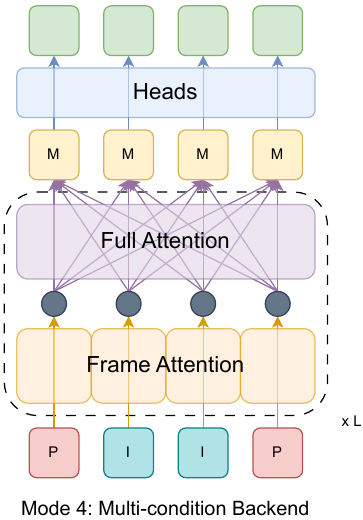}
        % \vspace{0.3em}
        % {\tiny\textbf{Mode 4: Multi-anchor Backend}}
    \end{minipage}
    % \vspace{-.5cm}
    \caption{Four training modes of \ours.
        \protect\adjustbox{valign=c}{\includegraphics[height=2\fontcharht\font`A]{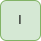}} and
        \protect\adjustbox{valign=c}{\includegraphics[height=2\fontcharht\font`A]{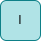}} represent the image tokens fed into the frontend and backend.
        \protect\adjustbox{valign=c}{\includegraphics[height=2\fontcharht\font`A]{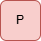}} represents the image tokens injected with pose.
        \protect\adjustbox{valign=c}{\includegraphics[height=2\fontcharht\font`A]{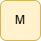}} represents the map tokens of a frame.
        \protect\adjustbox{valign=c}{\includegraphics[height=2\fontcharht\font`A]{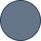}} and
        \protect\adjustbox{valign=c}{\includegraphics[height=2\fontcharht\font`A]{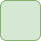}} represent the layer intermediate and final output.
        In each mode, \protect\adjustbox{valign=c}{\includegraphics[height=2\fontcharht\font`A]{figs/IF.drawio.png}},
        \protect\adjustbox{valign=c}{\includegraphics[height=2\fontcharht\font`A]{figs/IB.drawio.png}} and
        \protect\adjustbox{valign=c}{\includegraphics[height=2\fontcharht\font`A]{figs/P.drawio.png}} are fed into the transformer backbone $f_\theta$, with $L$ layers of frame attention and various inter-frame attentions.
        Pose and pointmap are regressed by the heads $f_{\psi}$.}
    \label{fig:training_modes}
    \vspace{-.3cm}

\end{figure*}
% FIXED: 1. mode1, mode2 字体不一致 [done]
%        2. check一下蓝色到底是不是maptoken [done]

As shown in Fig.~\ref{fig:training_modes}, the model is trained with four
shared-weight modes that differ only in attention masks and conditioning
patterns, matching test-time frontend, backend, and fine-stage inference.

\textbf{Training Frontend.}
Mode 1 trains online tracking and mapping: the first two frames initialize a
local coordinate system with full attention, and later frames use causal
inter-frame attention.
\begin{comment}
\(    \V F^{(1)}
    &=
    f_{\theta}^{\mathrm{fn}}
    \left(\V I_{1:N};\mathcal{A}_{1:N}^{\mathrm{causal}}\right).
\)
\end{comment}

\textbf{Training Backend without Memory Condition.}
Mode 2 trains backend refinement without memory condition. The
first half of the clip uses full attention, and the second half is decoded
causally from the refined prefix.
\begin{comment}
\(
    \V F^{(2)}
    &=
    f_{\theta}^{\mathrm{mix}}
    \left(
        \V I_{1:N};
        \mathcal{A}_{1:N/2}^{\mathrm{full}},
        \mathcal{A}_{N/2:N}^{\mathrm{causal}}
    \right).
\)
\end{comment}

\textbf{Training Backend with Memory Condition.}
Mode 3 trains backend refinement with memory condition.
Using detached Mode-2 predictions, we align a prefix span $[s,N/2)$ to an anchor
and encode it as pose-injected frames for the remaining target segment (with $s$ a random integer).
\begin{comment}

\(
\V F^{(3)}=
    f_{\theta}^{\mathrm{mix}}
    \left(
        \phi\left(\V I_{s:N/2}, \bar{\V g}^{(2)}_{s:N/2}\right),
        \V I_{N/2:N};
        \mathcal{A}_{s:t}^{\mathrm{full}},
        \mathcal{A}_{t:N}^{\mathrm{causal}}
    \right).
    \)
Here $t$ is randomly sampled in the second half of the clip, analogous to $s$ in the prefix, so that it separates the pose-condition prefix from the causal target segment, and $\phi$ is the pose-conditioning module.
\end{comment}

\textbf{Training Fine Stage with Memory Condition.}
Mode 4 matches the fine stage by conditioning the middle target chunk on front
and back neighboring condition chunks.

%% file: tex/experiments.tex
\section{Experiments}
% 1. setting 
% 1.1 training setup (data, hyper param, machine)
% 1.2 testing setup (data, metric) [ok]
% 2. small indoor （这里是跟vggt-slam是一个作用）
%   在7scene上可以提现fine的一个作用 
%   figure 1 出一个比较 [ok]
% 3. large outdoor [ok]
% 3.2 tracking performance (small-indoor, kitti, waymo)
%    table 1【ok】, table 2【ok】, 
% 3.3 reconstructino performance (waymo)
%    table 3【ok】， figure 2 (visual recon comparison with vggt-long) 【ok】
% overlap (long， （vkitti）)
% 4. ablation study
% 4.1 before and after fine (qualitative) 【ok】
% 5. test on the largest outdoor 【ok】

% comparison: SLAM-Former 224, long version [weicheng]

\begin{comment}
%TODO
% 1. outdoor 加上overlap试一下
% 2. table 1,2,3的横线，table需要确认一下 [done]
% 3. table 4 拆开。replica不要mast3r-slam 
% 4. 图5，grid现象
% 5. 大场景，解释说加上graph+posegraph [done]
% 6. 小场景误差小，不需要初始化，直接用frontend初始化，能优化回来 【因为vggt-slam的graph在replica不行】[done]
% 7. replica试一下没有coarse，只有fine[done]
\end{comment}

% Fixme:
% 1. kitti vis comparison between vggt-long vs ours
%       pick片段的结果 [done]
% 2. 17km vis comparison between vggt-long vs ours
%        可以pick一个小片段的结果

\subsection{Experimental Setup}

% \paragraph{Training setup.}
% Memory anchors provide local structural conditions, allowing the frontend to process frames scalably while the backend revisits long-range trajectory and geometry consistency.
% We use two domain-specific training configurations.
% The indoor configuration is trained on ARKitScenes, ScanNet++, ScanNet, HyperSim, BlendedMVS, and MegaDepth, using 12-frame clips and multi-scale inputs with long side 518.
% The outdoor configuration is trained on VirtualKITTI2, TartanGround, HabitatHM3D, ARKitScenes, ScanNet++, BlendedMVS, and MegaDepth, using 36-frame clips and multi-scale inputs with long side 224.
% Both configurations use a batch size of 1 per GPU, no gradient accumulation, weight decay 0.05, an initial learning rate of $1\times10^{-5}$, a minimum learning rate of $1\times10^{-8}$, 0.5 warm-up epochs, mixed-precision training, and 10 total epochs.
% Training is conducted on 48 A100 GPUs; the indoor run takes about 1 hour per epoch, while the outdoor run takes about 2.5 hours per epoch.
\paragraph{Training setup.}
% We train two domain-specific \ours variants to account for different scene scales and motion statistics.
We initialize two domain-specific \ours variants from a pretrained SLAM-Former model and further adapt them to different scene scales and motion statistics.
The indoor variant is trained with 12-frame clips at long side 518, and the outdoor variant with 36-frame clips at long side 224.
Both variants are trained for 10 epochs on 48 A100 GPUs.
See Appendix~\ref{app:training_details} for the dataset composition and optimization hyperparameters.
\begin{comment}
The dataset composition and optimization hyperparameters are provided in Appendix~\ref{app:training_details}.
    
\end{comment}

\paragraph{Evaluation protocol.}
We evaluate SLAMFormer-$\infty$\xspace on indoor benchmarks (Replica~\cite{sucar2021imap}, TUM RGB-D~\cite{sturm2012benchmark}, and 7-Scenes~\cite{glocker2013real}) and outdoor driving benchmarks (KITTI Odometry~\cite{geiger2012we} and Waymo Open Dataset~\cite{gaidon2016virtual}).
Tracking accuracy is reported as ATE RMSE in meters, and dense reconstruction is evaluated using pointmap accuracy, completeness, and Chamfer distance.
All reported metrics are lower-is-better.
Calibration-free methods are evaluated without camera intrinsics, while calibrated baselines use their standard calibrated inputs.
%For reproducibility, all indoor and outdoor benchmarks follow SLAM-Former and use fixed keyframe insertion and backend update intervals across datasets, without dataset-specific retuning.
% For KITTI long-sequence evaluation, the overlap is set to half of the chunk size, as reported in Table~\ref{table:kitti_ate}.

\subsection{Large Outdoor Scenes}

On outdoor sequences, the pose initialization for the iterative backend PGGO (fine stage) is obtained from a pose-graph optimization configured consistently with VGGT-Long~\cite{deng2025vggt}.

\paragraph{Tracking performance.}
% Long outdoor sequences stress SLAM systems with kilometer-scale motion, dynamic traffic, and wide-baseline observations.
% Table~\ref{table:kitti_ate} reports KITTI Odometry tracking results.
% SLAMFormer-$\infty$\xspace improves over VGGT-Long on full KITTI sequences, reducing the average ATE RMSE from 26.358 m to 23.011 m while remaining calibration-free and dense.
% This result suggests that the pose condition backend is beneficial when accumulated drift becomes significant.

\begin{table*}[!t]
  \centering
  \setlength{\tabcolsep}{3.0pt}
  \renewcommand{\arraystretch}{1.05}
  \resizebox{\linewidth}{!}{%
  \small
    \begin{tabular}{c|l|ccc|c|c|ccccccccccc}
    \toprule
        &\textbf{Method} & \textbf{LC} & \textbf{Calibration} & \textbf{Recon.} & \textbf{Avg.} & \textbf{Avg.$^*$} & \textbf{00} & \textbf{01} & \textbf{02} & \textbf{03} & \textbf{04} & \textbf{05} & \textbf{06} & \textbf{07} & \textbf{08} & \textbf{09} & \textbf{10}   \\ \midrule

        &\textitgray{seq. frames} & \textitgray{-} & \textitgray{-}&\textitgray{-}&	\textitgray{2109}&	\textitgray{2210} &	\textitgray{4542}&	\textitgray{1101}&	\textitgray{4661}&	\textitgray{801}&	\textitgray{271}&	\textitgray{2761}&	\textitgray{1101}&	\textitgray{1101}&	\textitgray{4071}&	\textitgray{1591}&	\textitgray{1201}\\
        
        &\textitgray{seq. length (m)} & \textitgray{-} & \textitgray{-}& \textitgray{-} & \textitgray{2012.243} & \textitgray{1968.147} & \textitgray{3724.19} & \textitgray{2453.20} & \textitgray{5067.23} & \textitgray{560.89} & \textitgray{393.65} & \textitgray{2205.58} & \textitgray{1232.88} & \textitgray{649.70} & \textitgray{3222.80} & \textitgray{1705.05} & \textitgray{919.52} \\
         	
        &\textitgray{seq. speed (m / frame)} & \textitgray{-} & \textitgray{-}& \textitgray{-} & \textitgray{0.95} & \textitgray{0.89} & \textitgray{0.82 } & \textitgray{\underline{2.23} } & \textitgray{1.09 } & \textitgray{0.70} & \textitgray{1.45 } & \textitgray{0.80} & \textitgray{1.12} & \textitgray{0.59} & \textitgray{0.79} & \textitgray{1.07} & \textitgray{0.77} \\
        
        &\textitgray{contains loop} & \textitgray{-} & \textitgray{-} & \textitgray{-}& \textitgray{-} & \textitgray{-} & \textitgray{\ding{51}} & \textitgray{\ding{55}} & \textitgray{\ding{51}} & \textitgray{\ding{55}} & \textitgray{\ding{55}} & \textitgray{\ding{51}} & \textitgray{\ding{51}} & \textitgray{\ding{51}} & \textitgray{\ding{55}} & \textitgray{\ding{51}} & \textitgray{\ding{55}} \\
        
        \midrule
        \multirow{3}{*}{\begin{turn}{90}Classic\end{turn}}&ORB-SLAM2~\cite{mur2017orb} (w/o LC) & \redtext{\ding{55}} & \redtext{\textit{Required}}& \redtext{\textit{Sparse}}  & 69.727 & 26.480 & 40.65  & 502.20  & 47.82  & \first{0.94}  & 1.30  & 29.95  & 40.82  & 16.04  & \third{43.09}  & 38.77  & \first{5.42}   \\ 
        
        & ORB-SLAM2\cite{mur2017orb} (w/ LC) & \greentext{\ding{51}} & \redtext{\textit{Required}}& \redtext{\textit{Sparse}}  & 54.816 & \first{9.464} & \first{6.03}  & 508.34  & \first{14.76}  & \second{1.02}  & 1.57  & \first{4.04}  & \second{11.16}  & \second{2.19}  & \second{38.85}  & \first{8.39}  & \second{6.63}   \\ 
        
        & LDSO~\cite{gao2018ldso} & \greentext{\ding{51}} & \redtext{\textit{Required}}& \redtext{\textit{Sparse}}  & \first{22.425} & 23.500  & 9.32  & \second{11.68}  & \third{31.98}  & 2.85  & 1.22  & \second{5.10}  & 13.55  & 2.96  & 129.02  & \third{21.64}  & 17.36   \\ 

        \midrule
        
        \multirow{10}{*}{\begin{turn}{90}Learning Based\end{turn}}&DROID-VO~\cite{teed2021droid} & \redtext{\ding{55}} & \redtext{\textit{Required}}& \greentext{\textit{Dense}}  & 54.188 & 51.187 & 98.43  & 84.20  & 108.80  & 2.58  & 0.93  & 59.27  & 64.40  & 24.20  & 64.55  & 71.80  & 16.91   \\ 
        
        &DPVO~\cite{teed2023deep}& \redtext{\ding{55}} & \redtext{\textit{Required}}& \redtext{\textit{Sparse}}  & 53.609 & 57.701 & 113.21  & 12.69  & 123.40  & \third{2.09}  & \first{0.68}  & 58.96  & 54.78  & 19.26  & 115.90  & 75.10  & \third{13.63}     \\ 
        
        &DROID-SLAM~\cite{teed2021droid} & - & \redtext{\textit{Required}}& \greentext{\textit{Dense}}  & 100.278 & 75.846  & 92.10  & 344.60  & 107.61 & 2.38  & 1.00  & 118.50  & 62.47  & 21.78  & 161.60  & 72.32 & 118.70   \\ 
        
        &DPV-SLAM~\cite{lipson2024deep} & \greentext{\ding{51}} & \redtext{\textit{Required}}& \redtext{\textit{Sparse}}  & 53.034 & 57.187  & 112.80  & \first{11.50}  & 123.53  & 2.50  & \third{0.81}  & 57.80  & 54.86  & 18.77  & 110.49  & 76.66  & 13.65   \\
        
        &DPV-SLAM++~\cite{lipson2024deep} & \greentext{\ding{51}} & \redtext{\textit{Required}}& \redtext{\textit{Sparse}}  & \third{25.749} & 27.138 & \third{8.30}  & \third{11.86}  & 39.64  & 2.50  & \second{0.78}  & \third{5.74}  & \third{11.60}  & \first{1.52}  & 110.90  & 76.70  & 13.70    \\ 

        \cmidrule(lr){2-18}
        
        % VGGT \cite{} & \redtext{\ding{55}} & \greentext{\textit{No}}  & \textitgray{3.56\%} & \textitgray{3.39\%}& \textitgray{1.65\%}& \textitgray{6.81\%}& \textitgray{6.81\%}& \textitgray{1.61\%}& \textitgray{9.36\%}& \textitgray{27.68\%}& \textitgray{2.72\%}& \textitgray{6.81\%}& \textitgray{1.84\%}& \textitgray{4.71\%} & \textitgray{6.24\%}   \\ 
        
        &MASt3R-SLAM~\cite{murai2025mast3r} & \greentext{\ding{51}} & \greentext{\textit{No Need}}& \greentext{\textit{Dense}} &	/ & /  & \textitgray{TL} &	\textitgray{\textitgray{TL}} & \textitgray{TL} & \textitgray{TL} & \textitgray{TL} & \textitgray{TL} & \textitgray{TL}  & \textitgray{TL} & \textitgray{TL} & \textitgray{TL} & \textitgray{TL}  \\ 
        &CUT3R~\cite{wang2025continuous} & \redtext{\ding{55}} & \greentext{\textit{No Need}}& \greentext{\textit{Dense}} &	/ & /  & \textitgray{OOM} &	\textitgray{OOM} & \textitgray{OOM} & 148.07 & 22.31 & \textitgray{OOM} & \textitgray{OOM}  & \textitgray{OOM} & \textitgray{OOM} & \textitgray{OOM} & \textitgray{OOM}  \\ 
        &Fast3R~\cite{yang2025fast3r} & \redtext{\ding{55}} & \greentext{\textit{No Need}}& \greentext{\textit{Dense}} &	/ & /  & \textitgray{OOM} &	\textitgray{OOM} & \textitgray{OOM} & \textitgray{OOM} & \textitgray{OOM} & \textitgray{OOM} & \textitgray{OOM}  & \textitgray{OOM} & \textitgray{OOM} & \textitgray{OOM} & \textitgray{OOM}  \\ 
        
        &VGGT~\cite{wang2025vggt} & \redtext{\ding{55}} & \greentext{\textit{No Need}}& \greentext{\textit{Dense}} &	/ & /  & \textitgray{OOM} &	\textitgray{OOM} & \textitgray{OOM} & \textitgray{OOM} & \textitgray{OOM} & \textitgray{OOM} & \textitgray{OOM}  & \textitgray{OOM} & \textitgray{OOM} & \textitgray{OOM} & \textitgray{OOM}  \\

        &VGGT-Long~\cite{deng2025vggt} & \greentext{\ding{51}} & \greentext{\textit{No Need}}& \greentext{\textit{Dense}} &	26.358 &	\third{19.298}&   \second{8.06}& 	96.96& 	34.16& 	6.83& 	4.16& 	9.15& 	\first{4.68}& 	\third{2.68}& 	63.15& 	32.24& 	27.87   \\ 
        \cmidrule(lr){2-18}
        &\textbf{SLAMFormer-$\infty$\xspace} & \greentext{\ding{51}} & \greentext{\textit{No Need}}& \greentext{\textit{Dense}} & \second{23.011} & \second{15.653} & 15.39 & 96.58 & \second{28.81} & 4.97 & 4.16 & 11.36 & 13.54 & 8.53 & \first{37.00} & \second{18.85} & 13.93   \\ 
        \bottomrule
        
    \end{tabular}
  }\vspace{-.1cm}
  \caption{KITTI Odometry tracking results. We report ATE RMSE [m] ($\downarrow$) on sequences 00--10. LC denotes loop closure, Avg.$^*$ excludes the high-speed Seq. 01, and OOM/TL denote CUDA out-of-memory on a single RTX 4090/tracking lost. Colors mark \first{first}, \second{second}, and \third{third} best results.}
  \label{table:kitti_ate}
\end{table*}

Table~\ref{table:kitti_ate} and Table~\ref{table:waymo_ate} report tracking results on KITTI and Waymo, respectively. On KITTI, SLAMFormer-$\infty$\xspace improves over VGGT-Long from $26.358$m to $23.011$m average ATE RMSE on full sequences, and on Waymo it further reduces the average ATE RMSE from $1.996$m to $1.813$m across urban segments with diverse speeds, lengths, and traffic densities. Taken together, these results indicate that the memory-conditioned frontend and backend improve tracking on large outdoor sequences by providing stronger long-range pose guidance and refinement than global pose alignment alone.

% Table~\ref{table:waymo_ate} evaluates tracking on Waymo segments with diverse speeds, lengths, and traffic densities.
% The same trend holds on Waymo, where SLAMFormer-$\infty$\xspace lowers the average ATE RMSE from 1.996 m for VGGT-Long to 1.813 m.
% These results support the proposed frontend/backend decomposition for scalable tracking and long-range refinement.

\begin{table*}[!t]
  \centering
  \setlength{\tabcolsep}{3.5pt}
  \renewcommand{\arraystretch}{1.05}
  \resizebox{\linewidth}{!}{%
  \small
\begin{tabular}{l|c|c|ccccccccc}
\toprule
\textbf{Segment ID}       & \textbf{Calib.}   & \textbf{Avg.} & \textbf{163453191} & \textbf{183829460} & \textbf{315615587} & \textbf{346181117} & \textbf{371159869} & \textbf{405841035} & \textbf{460417311} & \textbf{520018670} & \textbf{610454533} \\
\midrule
\textitgray{Frame num.}        & \textitgray{-}        & \textitgray{198}           & \textitgray{198}                & \textitgray{199}                & \textitgray{199}                & \textitgray{199}                & \textitgray{196}                & \textitgray{199}                & \textitgray{198}                & \textitgray{199}                & \textitgray{198}                \\
\textitgray{Segment length}   & \textitgray{-}        & \textitgray{172.533}       & \textitgray{159.963}            & \textitgray{42.301}             & \textitgray{165.149}            & \textitgray{351.213}            & \textitgray{272.661}            & \textitgray{85.743}             & \textitgray{265.906}            & \textitgray{134.552}            & \textitgray{62.739}             \\
\textitgray{Segment speed}    & \textitgray{-}        & \textitgray{0.871}         & \textitgray{0.808}              & \textitgray{0.213}              & \textitgray{0.830}              & \textitgray{1.765}              & \textitgray{1.391}              & \textitgray{0.431}              & \textitgray{1.343}             & \textitgray{0.676}              & \textitgray{0.317}              \\
\textitgray{Traffic}          & \textitgray{-}        & \textitgray{-}             & \textitgray{Low}                & \textitgray{High}               & \textitgray{Low}                & \textitgray{Low}                & \textitgray{Medium}                & \textitgray{Low}                & \textitgray{Medium}             & \textitgray{Low}                & \textitgray{High}               \\
\midrule
DROID-SLAM~\cite{teed2021droid}      & \redtext{\textit{Required}} & \third{4.396}         & \third{3.705}              & \first{0.301}              & \first{0.447}              & \third{8.653}              & 9.320              & 7.621              & \third{4.170}              & \textitgray{TL}                 & \first{0.264}              \\
MASt3R-SLAM~\cite{murai2025mast3r}     & \greentext{\textit{No Need}}  & 5.560         & 4.500              & \second{0.556}              & 1.833              & 12.544             & \third{8.601}              & \second{1.412}              & 5.428              & \third{7.910}              & 1.195              \\
CUT3R~\cite{wang2025continuous}           & \greentext{\textit{No Need}}  & 9.872         & 8.781              & 3.810              & 5.790              & 24.015             & 13.070             & 7.261              & 13.206             & 8.597              & 3.229              \\
Fast3R~\cite{yang2025fast3r}           & \greentext{\textit{No Need}}  & /             & \textitgray{OOM}                & \textitgray{OOM}                & \textitgray{OOM}                & \textitgray{OOM}                & \textitgray{OOM}                & \textitgray{OOM}                & \textitgray{OOM}                & \textitgray{OOM}                & \textitgray{OOM}                \\
VGGT~\cite{wang2025vggt}            & \greentext{\textit{No Need}}  & /             & \textitgray{OOM}                & \textitgray{OOM}                & \textitgray{OOM}                & \textitgray{OOM}                & \textitgray{OOM}                & \textitgray{OOM}                & \textitgray{OOM}                & \textitgray{OOM}                & \textitgray{OOM}                \\
VGGT-Long~\cite{deng2025vggt} & \greentext{\textit{No Need}}  & \second{1.996}         & \second{1.753}              & 2.629              & \second{0.559}              & \second{3.452}              & \first{3.343}              & \third{1.444}              & \first{1.541}              & \first{2.547}              & \third{0.455}      \\
\midrule
\textbf{SLAMFormer-$\infty$\xspace} & \greentext{\textit{No Need}}  & \first{1.813}         & \first{1.270}              & \third{0.616}              & \third{0.810}              & \first{1.464}              & \second{5.281}              & \first{0.568}              & \second{3.116}              & \second{2.825}              & \second{0.371}      \\
\bottomrule
\end{tabular}
  }  \vspace{-.2cm}
  \caption{Waymo tracking results. We report ATE RMSE [m] ($\downarrow$) on nine urban driving segments. Gray rows provide segment metadata, and OOM/TL denote CUDA out-of-memory/tracking lost. Colors mark \first{first}, \second{second}, and \third{third} best results.}
  \label{table:waymo_ate}
  \vspace{-.3cm}
\end{table*}

% \paragraph{Qualitative pose--geometry refinement.}
% Fig.~\ref{fig:kitti} shows that VGGT-Long mainly relies on global pose alignment, leaving the dense pointmaps largely unchanged and thus causing visible local geometric inconsistency. 
% By jointly refining poses and geometry, SLAMFormer-$\infty$ produces a more coherent large-scale reconstruction with better aligned local structures.

% \begin{figure}[!t]
%     \centering
%     \includegraphics[width=0.9\linewidth]{figs/kitti05.pdf}
%     \caption{Qualitative comparison on KITTI Seq. 05. VGGT-Long performs global pose alignment but leaves local geometry largely unrefined, while SLAMFormer-$\infty$ jointly optimizes pose and dense geometry, yielding a more coherent large-scale reconstruction.}
%     \label{fig:kitti}
%         % \vspace{-.5cm}
% \end{figure}

\paragraph{Reconstruction performance.}
Table~\ref{table:waymo_point} evaluates dense point-map reconstruction on Waymo.
Beyond trajectory accuracy, SLAMFormer-$\infty$\xspace also improves the average dense geometry over VGGT-Long: accuracy decreases from $1.182$ to $0.949$, completeness from $2.860$ to $2.777$, and Chamfer distance from $2.021$ to $1.863$.

Please also find in Fig.~\ref{fig:demo-long} the qualitative comparison on KITTI $05$ and $09$ sequences, where VGGT-Long's reconstructions are highly mismatched, while ours demonstrate smooth reconstructions.
Please find the Appendix~\ref{app:kitti_qualitative} for more demonstrations.
%where we show the qualitative reconstruction differences on KITTI.
%These results suggest that the tracking gains are accompanied by better dense geometry, rather than by pose improvement alone.

\begin{table*}[!t]
  \centering
  \setlength{\tabcolsep}{3.2pt}
  \renewcommand{\arraystretch}{1.05}
  \resizebox{\linewidth}{!}{%
  \small
\begin{tabular}{l|l|c|c|ccccccccc}
\toprule
\textbf{Segment ID}  & \textbf{Metric} & \textbf{Calib.}      & \textbf{Avg.} & \textbf{163453191} & \textbf{183829460} & \textbf{315615587} & \textbf{346181117} & \textbf{371159869} & \textbf{405841035} & \textbf{460417311} & \textbf{520018670} & \textbf{610454533} \\
\midrule
\textitgray{Frame num.}            & \textitgray{-}               & \textitgray{-}                    & \textitgray{198}           & \textitgray{198}                & \textitgray{199}                & \textitgray{199}                & \textitgray{199}                & \textitgray{196 }               & \textitgray{199}                & \textitgray{198}                & \textitgray{199}                & \textitgray{198}                \\
\textitgray{Segment length}       & \textitgray{-}               & \textitgray{-}                    & \textitgray{172.533}       & \textitgray{159.963}            & \textitgray{42.301}             & \textitgray{165.149}            & \textitgray{351.213}            & \textitgray{272.661}            & \textitgray{85.743}             & \textitgray{265.906}            & \textitgray{134.552}            & \textitgray{62.739}             \\
\textitgray{Segment speed}        & \textitgray{-}               & \textitgray{-}                    & \textitgray{0.871}         & \textitgray{0.808}              & \textitgray{0.213}              & \textitgray{0.830}              & \textitgray{1.765}              & \textitgray{1.391}              & \textitgray{0.431}              & \textitgray{1.343}              & \textitgray{0.676}              & \textitgray{0.317}              \\
\textitgray{Traffic}              & \textitgray{- }              & \textitgray{-}                    & \textitgray{-}             & \textitgray{Low}                & \textitgray{High}               & \textitgray{Low}                & \textitgray{Low}                & \textitgray{Medium}                & \textitgray{Low}                & \textitgray{Medium}             & \textitgray{Low}                & \textitgray{High}               \\
\midrule

\multirow{3}{*}{DROID-SLAM~\cite{teed2021droid}} & Accuracy $\downarrow$           & \multirow{3}{*}{\redtext{\textit{Required}}}             & \third{1.201}         & \second{0.781}              & \third{1.136}              & \third{2.247}              & \third{2.393}              & \first{1.090}              & \second{0.539}              & \first{0.740}              & \textitgray{TL}                 & \second{0.677}              \\
 & Completeness $\downarrow$           &  & \textitgray{8.540}         & 4.610              & 10.245             & 5.540              & 8.669              & 8.592              & 11.144             & 5.320              & \textitgray{TL}                 & 14.201             \\
 & Chamfer $\downarrow$        &  & \textitgray{4.870}         & 2.696              & 5.691              & 3.893              & 5.531              & 4.841              & 5.842              & \third{3.030}              & \textitgray{TL}                 & 7.439              \\
 \midrule
\multirow{3}{*}{MASt3R-SLAM~\cite{murai2025mast3r}}          & Accuracy $\downarrow$           & \multirow{3}{*}{\greentext{\textit{No Need}}}              & 3.772         & 3.189              & 2.988              & 3.787              & 4.689              & 4.436              & 1.166              & 4.637              & \third{6.417}              & 2.637              \\
 & Completeness $\downarrow$          &  & \third{3.177}         & \first{1.715}              & \second{3.284}              & \third{2.047}              & \second{2.981}              & \first{2.679}              & \first{2.895}              & \second{2.002}              & \third{4.429}              & \third{6.560}              \\
 & Chamfer $\downarrow$        &                      & \third{3.474}         & \third{2.452}              & \third{3.136}              & \third{2.917}              & \third{3.835}              & \third{3.558}              & \second{2.031}              & 3.319              & \third{5.423}              & \third{4.599}              \\
 \midrule
\multirow{3}{*}{CUT3R~\cite{wang2025continuous}}                & Accuracy $\downarrow$           & \multirow{3}{*}{\greentext{\textit{No Need}}}              & 3.884         & 3.580              & 1.144              & 2.418              & 3.712              & 3.679              & 4.346              & 2.012     & 12.320             & 1.744              \\
                     & Completeness $\downarrow$          &  & 6.801         & 8.251              & 9.352              & 8.748              & 8.537              & 5.467              & \third{3.393}              & 6.164     & \second{2.302}              & 8.999              \\
                     & Chamfer $\downarrow$        &  & 5.343         & 5.916              & 5.248              & 5.583              & 6.125              & 4.573              & 3.869              & 4.088     & 7.311              & 5.371              \\
                     \midrule
\multirow{3}{*}{VGGT-Long~\cite{deng2025vggt}}     & Accuracy $\downarrow$           & \multirow{3}{*}{\greentext{\textit{No Need}}}              & \second{1.182}         & \third{1.002}     & \first{0.395}     & \first{0.925}              & \second{1.668}              & \third{2.580}              & \third{0.679}              & \second{0.784}              & \second{1.358}              & \third{1.246}              \\
 & Completeness $\downarrow$          &  & \second{2.860}         & \third{2.762}     & \third{3.417}     & \first{1.738}              & \third{3.261}              & \second{2.791}              & \second{3.216}              & \first{1.840}              & 4.694              & \first{2.022}              \\
 & Chamfer $\downarrow$        &  & \second{2.021}         & \second{1.882}     & \second{1.906}     & \first{1.331}              & \second{2.465}              & \second{2.685}              & \first{1.948}              & \first{1.312}              & \second{3.026}              & \second{1.634}             \\
\midrule
\multirow{3}{*}{\textbf{SLAMFormer-$\infty$\xspace}} & Accuracy $\downarrow$ & \multirow{3}{*}{\greentext{\textit{No Need}}} & \first{0.949} & \first{0.601} & \second{0.467} & \second{1.454} & \first{1.064} & \second{1.964} & \first{0.251} & \third{1.268} & \first{0.935} & \first{0.540} \\
 & Completeness $\downarrow$ &  & \first{2.777} & \second{2.562} & \first{3.191} & \second{1.965} & \first{2.815} & \third{3.259} & 5.461 & \third{2.174} & \first{1.025} & \second{2.538} \\
 & Chamfer $\downarrow$ &  & \first{1.863} & \first{1.582} & \first{1.829} & \second{1.709} & \first{1.939} & \first{2.611} & \third{2.856} & \second{1.721} & \first{0.980} & \first{1.539}             \\
\bottomrule
\end{tabular}
  }  \vspace{-.2cm}
  \caption{Waymo point-map reconstruction results. We report accuracy, completeness, and Chamfer distance ($\downarrow$) against LiDAR point clouds; lower is better for all metrics. Gray rows provide segment metadata, and TL denotes tracking lost. Colors mark \first{first}, \second{second}, and \third{third} best results.}
  \label{table:waymo_point}
    \vspace{-.3cm}
\end{table*}

% The Waymo reconstruction numbers should be read together with Fig.~\ref{fig:waymo_recon_cmp}.
% The LiDAR ground truth is captured from a vehicle-mounted scanner and can miss high structures or camera-visible surfaces outside its vertical field of view.
% The visual comparison therefore complements the numeric metrics by showing whether the reconstructed geometry is globally aligned and free of duplicated structures.

\subsection{Small Indoor Scenes}

For indoor evaluation, we follow the VGGT-SLAM~\cite{maggio2026vggt} protocol by using a one-frame overlap between adjacent windows for error estimation and boundary-error correction.
Indoor benchmarks evaluate whether the pose condition design preserves local accuracy under short trajectories, narrow baselines, and frequent viewpoint changes.
Table~\ref{tab:indoor_tracking_reconstruction} reports aggregate tracking and reconstruction results on TUM RGB-D, 7-Scenes, and Replica.
SLAMFormer-$\infty$\xspace remains comparable to state-of-the-art indoor SLAM systems while preserving calibration-free dense reconstruction.
On 7-Scenes, it improves over VGGT-SLAM~\cite{maggio2026vggt} in both tracking and geometry: ATE RMSE decreases from $0.068$m to $0.046$m, and accuracy/completeness/Chamfer improve from $0.054/0.060/0.057$ to $0.029/0.049/0.039$. 
Similar trends hold on TUM RGB‑D and Replica, where our method achieves competitive tracking accuracy and high‑quality reconstruction.
%Similarly, consistent performance trends are observed on the TUM RGB-D and Replica datasets, where our method maintains highly competitive tracking accuracy and high-quality scene reconstruction.
% These results show that SLAMFormer-$\infty$\xspace retains strong indoor accuracy under the same calibration-free setting used outdoors.
In particular, matching-driven methods like MASt3R-SLAM~\cite{murai2025mast3r} and EC3R-SLAM~\cite{hu2025ec3r} excel on Replica due to near-perfect matching on noise-free simulated images, but degrade on real-world datasets.
SLAM-Former~\cite{yuan2025slam} achieves the overall best performance on indoor benchmarks, benefiting from its fully end-to-end learned architecture.
In contrast, \ours is designed primarily for global optimization over much longer trajectories, and its global optimization is not learned end-to-end.
Therefore, on short indoor benchmarks, ours may still underperform fully data‑driven models that are tailored specifically to the same distributions, whereas ours prioritizes global optimization for unbounded ones.
% These results show that SLAMFormer-$\infty$\xspace retains strong indoor accuracy under the same calibration-free setting used outdoors.

\begin{table*}[t]
  \centering
  \setlength{\tabcolsep}{3.2pt}
  \renewcommand{\arraystretch}{1.05}
  \resizebox{\linewidth}{!}{%
    \small
    \begin{tabular}{l c cccc ccc}
      \toprule
      \multirow{2}{*}{\textbf{Method}}
        & \multicolumn{1}{c}{\textbf{TUM RGB-D}}
        & \multicolumn{4}{c}{\textbf{7-Scenes}}
        & \multicolumn{3}{c}{\textbf{Replica}} \\
      \cmidrule(lr){2-2}
      \cmidrule(lr){3-6}
      \cmidrule(lr){7-9}
        & ATE$\downarrow$
        & ATE$\downarrow$
        & Acc.$\downarrow$
        & Comp.$\downarrow$
        & Chamf.$\downarrow$
        & ATE$\downarrow$
        & Acc.$\downarrow$
        & Comp.$\downarrow$ \\
      \midrule
     % MonoGS~\cite{matsuki2024gaussian}
     %   & 0.333 & 0.241 & -- & -- & -- & 0.456 & -- & -- \\
     % GO-SLAM~\cite{zhang2023go}
     %   & 0.171 & 0.099 & -- & -- & -- & 0.246 & -- & -- \\
     % GIORIE-SLAM~\cite{zhang2024glorie}
     %   & 0.155 & 0.094 & -- & -- & -- & 0.234 & 39.49 & 24.39 \\
     % PHOTO-SLAM~\cite{huang2024photo}
     %   & 0.414 & 0.191 & -- & -- & -- & 0.222 & -- & -- \\
     % Hi-SLAM2~\cite{zhang2025hi}
     %   & 1.750 & 0.104 & -- & -- & -- & 0.192 & 41.58 & 32.97 \\
     % ARTDECO~\cite{li2026artdeco}
     %   & 0.183 & 0.064 & -- & -- & -- & 0.166 & -- & -- \\
      CUT3R~\cite{wang2025continuous}
        & 0.113 & 0.073 & 0.032 & 0.047 & 0.040 & 0.170 & 7.52 & 3.62 \\
      StreamVGGT~\cite{zhuo2025streaming}
        & 0.187 & 0.081 & 0.058 & 0.057 & 0.057 & 0.125 & 9.88 & 4.73 \\
      VGGT-SLAM~\cite{maggio2026vggt}
        & 0.084 & 0.068 & 0.054 & 0.060 & 0.057 & 0.071 & 7.52 & 5.86 \\
      MASt3R-SLAM~\cite{murai2025mast3r}
        & 0.061 & 0.065 & 0.065 & 0.067 & 0.056 & 0.045 & 2.92 & 2.25 \\
      EC3R-SLAM~\cite{hu2025ec3r}
        & 0.070 & 0.075 & 0.025 & 0.054 & 0.040 & 0.041 & 2.82 & 2.07 \\
      %ViSTA-SLAM~\cite{zhang2025vista}
      %  & 0.052 & 0.056 & 0.051 & 0.055 & 0.045 & 0.108 & 11.44 & 5.34 \\
      VGGT-Long~\cite{deng2025vggt}$^{+}$
        & 0.082 & 0.089 & 0.049 & 0.081 & 0.065 & 0.071 & 5.80 & 3.17 \\
      SLAM-Former~\cite{yuan2025slam}
        & \textbf{0.039}
        & \textbf{0.042} & \textbf{0.017} & \textbf{0.037} & \textbf{0.027}
        & \textbf{0.030} & \textbf{2.09} & \textbf{1.56} \\
      \midrule
      SLAMFormer-$\infty$\xspace
        & 0.066 & 0.046 & 0.029 & 0.049 & 0.039 & 0.052 & 6.00 & 3.33 \\
      SLAMFormer-$\infty$\xspace (w/o fine)
        & 0.068 & 0.047 & 0.030 & 0.049 & 0.040 & 0.061 & 6.11 & 3.39 \\
      \bottomrule
    \end{tabular}%
  }  
  \caption{
    Indoor tracking and reconstruction results on TUM RGB-D, 7-Scenes, and Replica.
    We report ATE RMSE [m] for tracking, and accuracy, completeness, and Chamfer distance for reconstruction.
    Lower is better for all metrics.
    $^{+}$ indicates results from our own run.
  }
  \label{tab:indoor_tracking_reconstruction}
  \vspace{-.3cm}
\end{table*}
% FIXME： 1. explain why mast3r-slam's matching method is better.解释一下。
%         2. 看看测一下vggt-long的 跑一下
% 3. 调换一下数据集的顺序 7-scenes

% The fine stage provides consistent but modest gains on the indoor benchmarks; its effect is analyzed in the ablation study.

\subsection{Ablation Study: Before and After Fine Stage (PGGO)}
\begin{figure}[!t]
    \centering
    \includegraphics[width=1\linewidth]{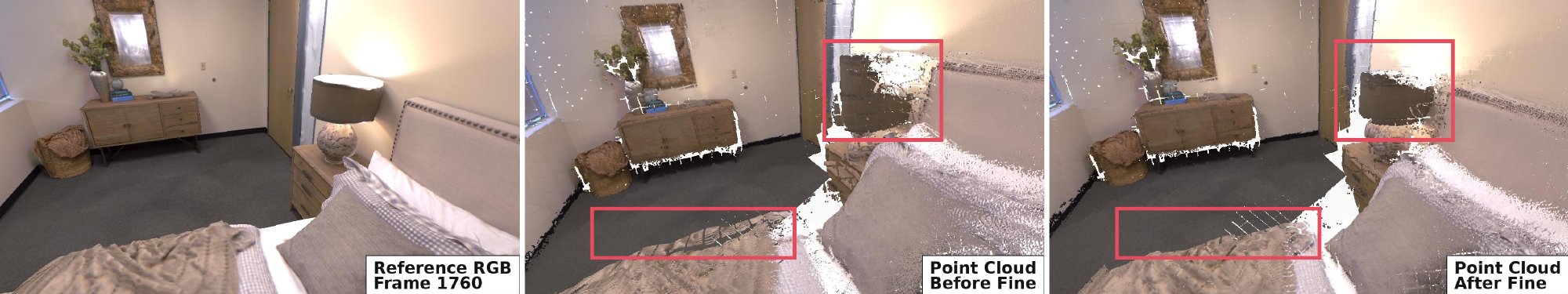}  
    \caption{Qualitative effect of the fine stage (PGGO) on Replica. Compared with the coarse prediction, the fine-stage output produces cleaner local surfaces and more stable alignment.}
    \label{fig:fine}
    \vspace{-.3cm}
\end{figure}

Table~\ref{tab:indoor_tracking_reconstruction} also compares SLAMFormer-$\infty$\xspace with and without the fine stage.
Quantitatively, the fine stage brings consistent gains across the indoor benchmarks.
The clearest improvement appears on Replica, where ATE RMSE decreases from $0.061$m to $0.052$m and reconstruction accuracy/completeness improve from $6.11/3.39$ to $6.00/3.33$.
TUM RGB-D and 7-Scenes show the same overall trend, indicating that the fine stage improves local geometric consistency, although the numerical margins remain limited.
By contrast, the qualitative effect is more evident: Fig.~\ref{fig:fine} shows cleaner point-map surfaces and reduced local drift after the fine stage, making the visual improvement more noticeable than the score differences alone suggest.

